\documentstyle[apalike,11pt]{article}

\input{epsf}

\begin{document}

\title{Ensembles of Radial Basis Function Networks for Spectroscopic 
Detection of Cervical Pre-Cancer}

\author{Kagan Tumer\footnote{NASA Ames Research Center/Caelum Research; {\tt kagan@ptolemy.arc.nasa.gov}}, 
Nirmala Ramanujam\footnote{Dept. Biochemistry and Biophysics, Johnson Research Foundation, University of Pennsylvania, School of Medicine; {\tt nimmi@mail.med.upenn.edu}}, 
Joydeep Ghosh\footnote{Dept. of Electrical and Computer Engr., The University of Texas at Austin; {\tt ghosh@ece.utexas.edu}}, 
and Rebecca Richards-Kortum\footnote{Biomedical Engineering Program, The University of Texas at Austin; {\tt kortum@mail.utexas.edu}}
}

\date{}

\maketitle

\begin{abstract}
The  mortality  related  to
cervical  cancer can be substantially reduced through early 
detection and treatment.
However, current detection techniques, such as Pap smear and
colposcopy, fail to achieve a concurrently high sensitivity and 
specificity. 
{\em In  vivo} fluorescence spectroscopy is a technique which
quickly,   {\em non-invasively}   and
quantitatively  probes  the  biochemical  and   morphological
changes  that  occur  in pre-cancerous tissue.
A multivariate statistical algorithm was used to extract clinically 
useful information from tissue spectra acquired from 361 cervical
sites from 95 patients at 337, 380 and 460 nm excitation wavelengths.
The multivariate statistical analysis was also employed to reduce the 
number of fluorescence excitation-emission wavelength pairs required
to discriminate healthy tissue samples from pre-cancerous tissue samples.
The use of connectionist methods such as multi layered perceptrons, 
radial basis function networks, and ensembles of such networks was
investigated.
RBF ensemble algorithms based on fluorescence spectra potentially 
provide automated, and  near real-time  implementation  of pre-cancer  
detection  in  the hands   of   non-experts.
The results are more reliable, direct and accurate than those
achieved by either human experts or multivariate statistical
algorithms.
\end{abstract}

\section{Introduction}
Cervical carcinoma is the second most common cancer in
women  worldwide, exceeded only by breast cancer~\cite{acs95}.
The  mortality  related  to cervical  cancer can be reduced if this 
disease is  detected at its pre-cancerous
state, known as squamous intraepithelial lesion (SIL)~\cite{wrku94}. 
Even though widespread use of organized screening (Pap smear) and
diagnostic (colposcopy) programs are currently in place, approximately
15,900 new cases of cervical cancer and 4,900 cervical cancer related
deaths were reported in the United States alone, in 1995~\cite{acs95}.

Currently, the primary screening tool for the detection of
cervical cancer and its precursor is the Pap smear~\cite{kuhe94}. 
In a Pap test, a large number  of
cells  obtained  by  scraping  the  cervical  epithelium  are
smeared  onto  a slide which is then fixed and  stained  for
cytologic examination. 
Each smear is then examined under a microscope for the presence 
of neoplastic cells~\cite{who88}.
The Pap smear is unable to achieve a concurrently
high sensitivity\footnote{Sensitivity is the correct classification
percentage on the pre-cancerous tissue samples.}
and high specificity\footnote{Specificity is the correct classification
percentage on normal tissue samples.}~\cite{fair95}.
The accuracy of the Pap
smear is limited by both sampling and reading errors~\cite{wilk90}.
Approximately
60\% of false-negative smears are attributed to insufficient sampling;
the remaining 40\% are due to reading errors. Because of the monotony
and fatigue associated with reading Pap smears (50,000-300,000 cells
per slide), the American Society of Cytology has proposed that a
cyto-technologist should be limited to evaluating  no more than
12,000 smears annually~\cite{koss89}. As a result, accurate Pap smear
screening is labor intensive and requires highly trained professionals.
Some new tools (Thinkprep, Papnet, autopap) to assist cyto-technologists 
have recently been introduced, but they are all based on 
invasive techniques~\cite{korm96}. 

A  patient  with a Pap smear interpreted as indicating
the presence of SIL is followed up by a diagnostic procedure
called colposcopy~\cite{kuhe94}. 
During a colposcopic examination, the
cervix is stained with acetic acid and viewed through a low power
microscope to identify potential pre-cancerous sites. 
Subsequently, suspicious sites
are biopsied and then histologically examined to confirm the presence,
extent and severity of the lesion~\cite{budu91}.
Colposcopic
examination in expert hands maintains a high sensitivity  at
the  expense  of  a  significantly low specificity, leading
to many unnecessary biopsies~\cite{mitc94}.
In spite of the poor specificity  of
this  technique, extensive training is required  to  achieve
this  skill  level. Furthermore, since this procedure
involves biopsy, which
requires histologic evaluation, diagnosis is not immediate~\cite{kuhe94}. 

Laser induced fluorescence spectroscopy is an optical technique 
which has   the   capability   to  quickly,   non-invasively   and
quantitatively  probe  the  biochemical  and   morphological
changes  that  occur  as  tissue  becomes  neoplastic.   
The altered biochemical
and morphological state of the neoplastic tissue is reflected 
in the spectral characteristics of the measured fluorescence. 
This  spectral information can be correlated to tissue 
histo-pathology, the current ``gold standard'' to develop 
clinically effective screening algorithms.
These mathematical algorithms can be implemented in software,
potentially enabling automated, fast, non-invasive and accurate
pre-cancer detection in hands of non-experts.
Although a complete understanding of the quantitative 
information contained within a tissue fluorescence spectrum 
has not been achieved, many groups have investigated
the use of fluorescence 
spectroscopy for real-time, non-invasive, automated 
characterization of tissue 
pathology~\cite{brwa95,cori90,lomu89,mabr92,rira91,scfr92,yuya87}.

A detection technique for human cervical
pre-cancer  based on laser induced fluorescence spectroscopy
has  been  developed  recently~\cite{rami96}.
Discrimination  was achieved using a  multivariate
statistical  algorithm  (MSA) based on  Principal  Component
Analysis (PCA) and  Logistic  Discrimination  of  tissue  spectra
acquired   {\em in   vivo}.  
This  linear  method   of   algorithm
development demonstrated an improved classification accuracy
relative  to  both  the Pap smear and  colposcopy  in
expert  hands. In this article, we investigate neural 
network based non-linear methods for algorithm development, 
and compare them 
to both the MSA and conventional clinical methods. Specifically, we
investigate the performance of Multi-Layered Perceptron (MLP) 
and Radial Basis Function (RBF) networks, and ensembles of 
these networks, on cervical tissue fluorescence spectra. 
The connectionist methods aim at improving the 
classification accuracy and reliability of the MSA, as well as
simplifying the decision making process.
Section~\ref{sec:data} presents the data collection/processing
techniques. In Section~\ref{sec:algo}, the MSA, and the neural 
network based methods are described. Section~\ref{sec:resu}
contains the results of our analysis and compares the neural network
results to that of the MSA  and current clinical
detection methods.
A discussion of the results is given in Section~\ref{sec:disc}. 

\section{Data Collection and Processing}
\label{sec:data}
\subsection{Instrumentation and Clinical Measurements}
\label{sec:clin}
A portable fluorimeter consisting of two nitrogen pumped-dye lasers, 
a fiber-optic probe and a polychromator coupled to an intensified diode 
array controlled by an optical multi-channel analyzer was utilized to 
measure fluorescence spectra from the cervix {\em in vivo} at three excitation 
wavelengths: 337, 380 and 460 nm \cite{rami96}. 
Data acquisition, calibration and  processing have been described in
detail elsewhere~\cite{rami96}.

A randomly selected group of non-pregnant patients referred to 
the colposcopy clinic of the University of Texas MD Anderson Cancer Center 
on the basis of abnormal cervical cytology was asked to participate in the 
{\em in vivo} fluorescence spectroscopy study. 
Informed consent was  obtained from
each patient who participated and the study was reviewed and approved by 
the Institutional Review Boards of the University of Texas, Austin and the
University of Texas, MD Anderson Cancer Center.  

Each patient underwent a
complete history and a physical examination including a pelvic exam, a 
Pap smear and colposcopy of the cervix, vagina and vulva. 
After colposcopic examination of the cervix, but before tissue
biopsy, fluorescence spectra were acquired on average from two 
colposcopically abnormal sites, two colposcopically normal squamous 
sites and 1 normal columnar site (if colposcopically visible) from each 
patient, from a total of 361 cervical
sites in 95 patients.

\begin{table}[ht]
\caption{Histo-pathologic classification of samples
from the training   and  test  sets.
Note,   biopsies   for
histological    evaluation   were    not    obtained    from
colposcopically normal squamous and columnar tissue sites to
comply with routine patient care procedure.}
\label{tab:data}
\begin{center}
\begin{tabular}{cc|cc} \hline \hline
         Histo-pathology  & Target Classification &   Training  Set & Test Set \\ \hline
         Normal Squamous  &  &      94     &  94  \\
         Normal Columnar  &  non-SIL &      13     &  14  \\
          Inflammation    &  &      15     &  14  \\ \hline
             LG SIL       &  SIL &      23     &  24  \\
             HG SIL       &  &      35     &  35  \\ \hline
\end{tabular}
\end{center}
\end{table}

Tissue biopsies were obtained only from abnormal sites 
identified by colposcopy and subsequently analyzed by the probe to 
comply with routine patient care procedure. 
All tissue biopsies were 
fixed in formalin and submitted for histologic examination. 
Hemotoxylin 
and eosin stained sections of each biopsy specimen were evaluated by a 
panel of four board certified pathologists and a consensus diagnosis was 
established using the Bethesda classification system~\cite{wrku94}.
This classification 
system which has previously been used to grade cytologic specimens has 
now been extended to classification of histology samples. 
Samples were classified as normal squamous (NS), normal columnar (NC), 
inflammation\footnote{In this paper we will not focus on the
classification of tissues with inflammation. Evaluation of these tissues
using both MSA and neural networks indicates that they are nearly 
indistinguishable from SILs based on the spectral data presented 
here~\cite{rami95a,rami95b}. To remedy the situation, different optical
spectroscopic techniques are needed.},
low grade (LG) SIL and high grade (HG) SIL. 
Table~\ref{tab:data} provides the
number of samples in the training (calibration) and test (prediction) sets.

\subsection{Spectral Data}
\begin{figure}[htb]
\centerline{\epsfbox{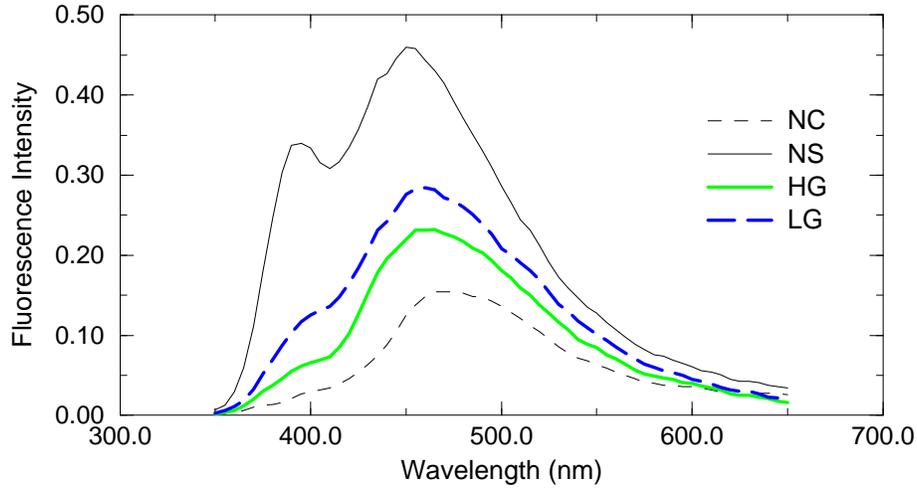}}
  \caption{Fluorecsence spectra from a typical patient at  337 nm excitation.}
  \label{fig:flu337}
\vspace*{.1in}
\end{figure}

Figure~\ref{fig:flu337} illustrates average fluorescence spectra 
per site acquired from cervical sites at 337 nm excitation from a typical
patient.
All fluorescence intensities are reported in the same set of 
calibrated units. 
Evaluation of the spectra at 337 nm excitation indicates
that the fluorescence intensity of SILs (LG and HG)
is less than that of the 
corresponding normal squamous tissue; however, their fluorescence
intensity is greater than that of the 
corresponding normal columnar tissue over the entire emission 
spectrum.

\begin{figure}[htb]
\centerline{\epsfbox{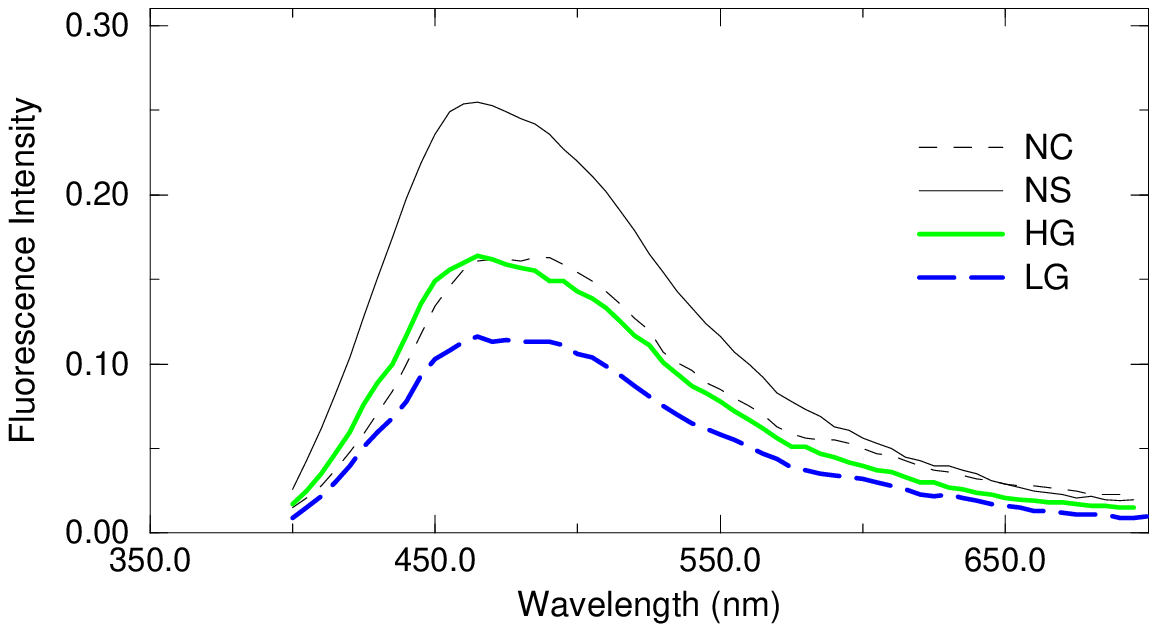}}
  \caption{Fluorecsence spectra from a typical patient at  380 nm excitation.}
  \label{fig:flu380}
\vspace*{.1in}
\end{figure}

\begin{figure}[htb]
\centerline{\epsfbox{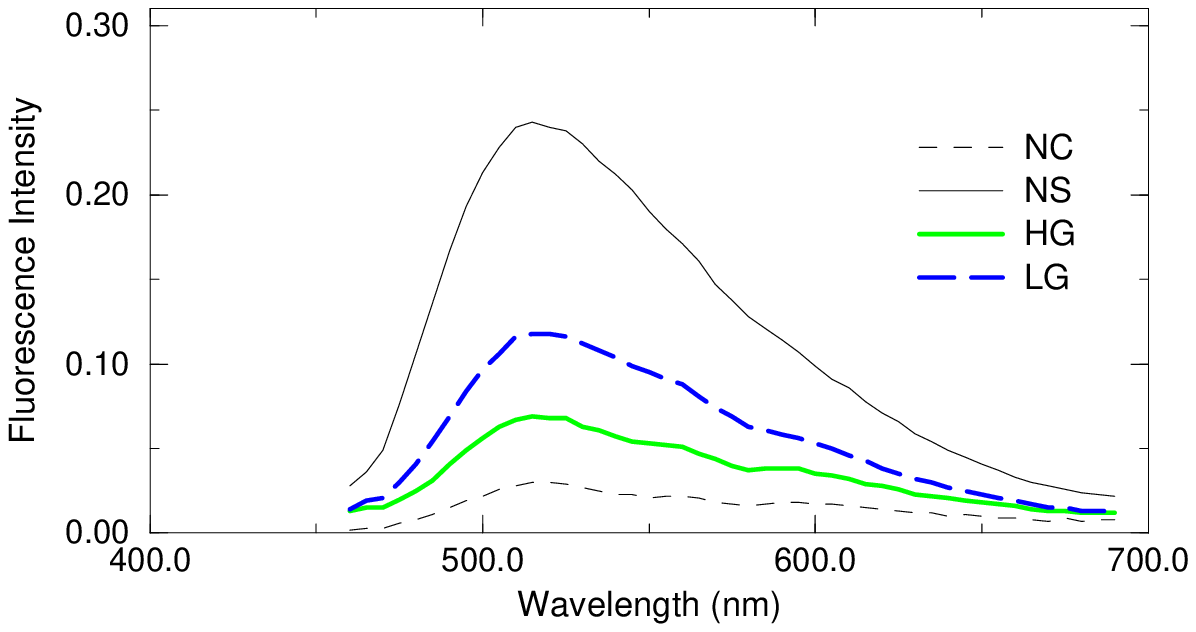}}
  \caption{Fluorecsence spectra from a typical patient at  460 nm excitation.}
  \label{fig:flu460}
\vspace*{.1in}
\end{figure}

Figure~\ref{fig:flu380} illustrates average fluorescence spectra per 
site acquired from cervical sites at 380 nm excitation, from the same 
patient. In Figure~\ref{fig:flu380}, the fluorescence intensity of SILs 
is less than that of the corresponding normal squamous tissue, with the 
LG SIL exhibiting the weakest fluorescence intensity over the entire 
emission spectrum. Note that the fluorescence intensity of the normal 
columnar sample is indistinguishable from that of the HG SIL. 
Figure~\ref{fig:flu460} illustrates spectra at 460 nm excitation from 
the same patient. Evaluation of Figure~\ref{fig:flu460} 
indicates that the fluorescence 
intensity of SILs is less than that of the corresponding normal squamous 
tissue and greater than that of the corresponding normal columnar sample 
over the entire emission spectrum. 

Tissue fluorescence spectra at 337 nm 
excitation consists of intensities at a total of 59 emission wavelengths; 
tissue spectra at 380 nm excitation consists of intensities at a total of 
56 emission wavelengths and that at 460 nm excitation consists of 
intensities at 45 emission wavelengths. Hence, 
fluorescence spectra at all three excitation 
wavelengths comprise of fluorescence intensities at a total of 
160 excitation-emission wavelengths pairs.

\section{Algorithm Development}
\label{sec:algo}
In this section, the development of the multivariate statistical 
algorithm and the neural network based algorithms are described.
Each type of algorithm was utilized to develop a detection method
that can effectively discriminate between SILs and non-SILs
(normal squamous and normal columnar).

\subsection{Multivariate Statistical Algorithms}
\label{sec:msa}
\subsubsection{Full-Parameter Multivariate Statistical Algorithm}
\label{sec:fulpar}
The Multivariate Statistical Algorithm (MSA) development described 
in~\cite{rami96} consists of the following five steps:
\begin{enumerate}
\item{} pre-processing to reduce inter-patient and intra-patient 
variation of fluorescence spectra from a tissue type,
 using either normalization 
or normalization followed by mean-scaling, 
\item{} dimension reduction of the pre-processed tissue fluorescence
spectra using Principal Component Analysis (PCA), 
\item{} selection of diagnostically relevant principal components, 
using an unpaired, one-sided student's t-test, 
\item{} development of a classification algorithm based on logistic 
discrimination, 
using the diagnostically relevant principal components, 
and finally
\item{} retrospective and prospective evaluation of the algorithm's 
accuracy on a training and test set, respectively. 
\end{enumerate}

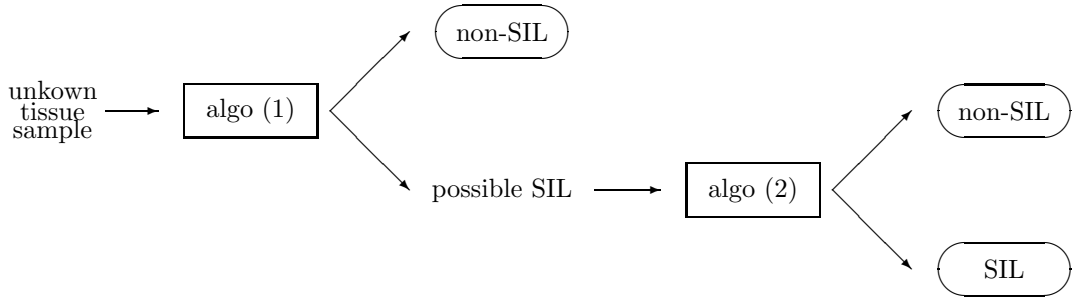
\begin{figure}
\begin{picture}(400,140)(-30,0)

{ \small

    \put(0,88){\makebox(0,0){unkown}}
    \put(0,80){\makebox(0,0){tissue}}
    \put(0,72){\makebox(0,0){sample}}

    \put(20,80){\vector(1,0){20}}

    \put(75,80){\makebox(0,0){algo (1)}}
    \put(50,70){\framebox(50,20)}

    \put(105,80){\vector(1,1){30}}
    \put(105,80){\vector(1,-1){30}}

    \put(170,110){\makebox(0,0){non-SIL}}
    \put(170,110){\oval(50,20)}

    \put(170,50){\makebox(0,0){possible SIL}}

    \put(205,50){\vector(1,0){25}}

    \put(265,50){\makebox(0,0){algo (2)}}
    \put(240,40){\framebox(50,20)}

    \put(295,50){\vector(1,1){30}}
    \put(295,50){\vector(1,-1){30}}

    \put(360,80){\makebox(0,0){non-SIL}}
    \put(360,80){\oval(50,20)}

    \put(360,20){\makebox(0,0){SIL}}
    \put(360,20){\oval(50,20)}
}
\end{picture}
\caption{Schematic representation of the composite MSA algorithm.}
\label{fig:schema}
\end{figure}

This process of algorithm development was applied to tissue 
fluorescence spectra acquired at all three excitation wavelengths, as 
described in detail in~\cite{rami96}.
Discrimination between the SILs and the two normal tissue types
was achieved using a composite algorithm of two independently developed
{\em constituent} algorithms. Constituent algorithm (1), which is 
based on tissue spectra that have been pre-processed by 
normalization, discriminates between SILs and normal squamous 
tissue samples. Constituent algorithm (2), which is based on
tissue spectra that have been pre-processed by normalization
followed by mean-scaling, discriminate between SILs and normal
columnar tissue samples. The classification outputs from both
constituent algorithms were used to determine whether a sample
being evaluated is SIL/non-SIL. A sample is first presented
to constituent algorithm (1). If it is classified as  non-SIL,
the algorithm terminates. If it is classified as SIL, then
the sample is presented to constituent algorithm (2), and 
the result of that algorithm determines the final classification
of the tissue sample. Figure~\ref{fig:schema} illustrates this procedure.

\subsubsection{Reduced-Parameter Multivariate Statistical Algorithms}
\label{sec:redpar}
The full-parameter MSA utilizes fluorescence spectra
at all three excitation wavelengths to develop a classification scheme
for cervical pre-cancer detection; the fluorescence spectra at these three
excitation wavelengths correspond to fluorescence intensities at a total
of 160 excitation-emission wavelength pairs (5 nm spectral resolution).
However, there is a significant cost penalty for using all 160 values.
To alleviate this concern, a more cost-effective 
fluorescence imaging system can be developed
if the number of required excitation-emission wavelength pairs at which
fluorescence intensities need to be recorded is significantly reduced.
For example, if the number of required excitation-emission wavelength
pairs that need to measured can be reduced by an order of magnitude,
the polychromator and intensified diode array can be replaced by a 
mechanical filter assembly and a single channel detector. The resulting
system represents a substantial decrease in cost and complexity of 
this instrumentation.

We have shown that component loadings calculated from Principal Component
Analysis~\cite{rami96} can be used to reduce the number of fluorescence
excitation-emission wavelength pairs required to generate 
the constituent algorithms from
160 to 13, with a minimal decrease in classification accuracy.
The component loadings represent
the correlation between a principal component and the pre-processed
fluorescence spectra at each excitation-emission wavelength pair.
More precisely, the diagnostically relevant principal components of
the full-parameter set (160 variables) selected using the students 
t-test are used to calculate the component loadings.
The intensity/wavelength pairs that show a strong correlation 
to the relevant principal components form the reduced-parameter
set.
Table~\ref{tab:flu} shows
the fluorescence intensities at the reduced number of excitation-emission
wavelength pairs. These pairs are used to redevelop constituent 
algorithms (1) and (2) using the MSA process.

\begin{table}[ht]
\caption{ Fluorescence intensities at 13 excitation-emission
wavelength  pairs needed to re-develop the two reduced-parameter
constituent algorithms. Aside from the pre-processing, the 
two sets only differed in two selections, identified by $\ast$.}
\label{tab:flu}
\begin{center}
\begin{tabular}{ccc} \hline \hline
  Features for Algorithm (1)   &    Features for Algorithm (2)   \\ 
  (Normalized)&    (Normalized and Mean Scaled)\\ 
 $\lambda_{ex}$, $\lambda_{em}$  & $\lambda_{ex}$, $\lambda_{em}$  \\ \hline
   337, 410 nm   &    337, 410 nm   \\ 
   337, 430 nm   &    337, 430 nm   \\ 
   337, 510 nm   &    337, 510 nm   \\ 
   337, 580 nm   &    337, 580 nm   \\ 
   380, 410 nm   &    380, 410 nm   \\ 
   380, 430 nm   &    380, 430 nm   \\ 
   380, 510 nm   &    380, 510 nm   \\ 
   380, 580 nm   &    380, 580 nm   \\ 
   380, 640 nm   &    \hspace*{.1in} 380, 600 nm $\ast$\\ 
   460, 580 nm   &    460, 580 nm   \\ 
   460, 600 nm   &    460, 600 nm   \\ 
   460, 620 nm   &    460, 620 nm   \\ 
   460, 640 nm   &    \hspace*{.1in} 460, 660 nm $\ast$ \\ \hline 
\end{tabular}
\end{center}
\end{table}

\subsection{Algorithms Based on Neural Networks}
\label{sec:ann}
The second stage of algorithm development consists of
evaluating the applicability of neural networks to this
problem. For this study, we consider two types of feed 
forward neural networks, the Multi-Layered Perceptron (MLP) 
and the Radial Basis function (RBF) network~\cite{tura97}. 

\subsubsection{Multi-Layered Perceptrons}
The MLP, probably the most commonly used neural network
architecture, is a feed-forward neural network 
with an input layer, an output layer and possibly multiple
hidden layers. Each layer is connected only to the 
subsequent layer by variable weights which are adjusted
to minimize a predetermined cost function, such as the Mean
Square Error (MSE). 
For an MLP with one hidden layer, the
responses of the $k$th hidden unit, $h_k$, and the
$j$th output, $o_j$, are respectively given by:
\begin{eqnarray}
h_k = g(\sum_i v_{ki} x_i);
\label{eq:hid}
\end{eqnarray}
and
\begin{eqnarray}
o_j = f(\sum_k w_{jk} h_k),
\label{eq:out}
\end{eqnarray}
where the input to hidden connections strength are given by $v$,
the hidden to output connection strengths are given by $w$,
and indices $i$ and $k$ sum over the input ($x$) and hidden
units respectively. The activation functions $f(\cdot)$ and
$g(\cdot)$ are sigmoidal functions.

In order to adapt the weights, the {\em backpropagation} 
algorithm is generally used. The principle of this algorithm
is based on distributing the ``blame'' or the contribution of 
each unit to the overall error, across the proper weights. 
Further details can be found in \cite{hayk94,bish95}. In this study
we only explore MLPs with a single hidden layer.

\subsubsection{Radial Basis Function Networks}
Radial Basis Function networks are feed-forward networks
with a single hidden layer where the activation function
is a radially symmetric basis function. 
The output units perform a weighted sum over the outputs
of the hidden units (also called kernels). 
One class of radial basis functions that is of particular
interest consists of those with Gaussian kernels, or where the 
basis', ($R_j({\bf x})$), have the following activation 
function~\cite{hayk94}:
\begin{eqnarray}
R_{j}({\bf x}) = e^{- \frac{1}{2}\frac{\|{\bf x}-{\bf x_{j}}\|^2}{\sigma_j^2}} ,
\label{eq:rbf}
\end{eqnarray}
where $\sigma_j$ determines the width of the receptive field, and
${\bf x_j}$ determines the centroid of the $j$th kernel, respectively.
The $j$th hidden node has a maximum output of 1 when input 
${\bf x}={\bf x_{j}}$.  
Important parameters in the design of RBF networks
include the number, location and receptive field widths of the kernels. 

\subsubsection{Combining Multiple Networks}
The performance of a given MLP or RBF network depends on many
parameters, including size, learning rate, training strategy
and initial weights. These differences result in different
classification decisions, making the selection of a single
``best'' network a delicate matter. 
This problem is further
compounded when the amount of training patterns is limited,
because the definition of ``best''  depends on
the particular validation set chosen. In such cases, it
is difficult to ascertain
whether one network will outperform all others given a different test
set, as the validation sets are small.
Furthermore, selecting only one classifier discards a large amount of
potentially relevant information. In order to use all the available
data, and to increase both the performance and the reliability of the 
methods, one can pool the outputs of the individual classifiers 
before a classification decision is made 
\cite{hawa92,hasa90,peco93,tugh96,wolp92}.

In this study we use the median
combiner, $f_i^{med}$, which belongs to the class of order statistics 
combiners introduced in~\cite{tugh95f}, and
the averaging combiner, $f_i^{ave}$ which performs
an arithmetic average of the corresponding outputs:
\begin{eqnarray}
f_i^{med}(x)\; & = & \; \left\{ \begin{array}{ll}
     \frac {f^{\frac{N}{2}:N}_i(x) \; + \:
f^{\frac{N}{2}+1:N}_i(x)}{2} & \mbox{if $N$ is even} \\ 
f^{\frac{N+1}{2}:N}_i(x)           & \mbox{if $N$ is odd},
                       \end{array} \right. 
\label{eq:med}
\end{eqnarray}
and
\begin{eqnarray}
f_i^{ave}(x) = \frac{1}{N} \; \sum_{m=1}^N f_i^m(x),
\label{eq:ave}
\end{eqnarray}
where $N$ represents the number of available classifiers, and
$f_i^{m}(x)$ the $i$th output of the $m$th classifier for input $x$.
We selected these two combiners, because of their 
simplicity in both interpretation and implementation, and because 
they typically result in good and robust performance~\cite{kiha96,tugh95f}.

\section{Results}
\label{sec:resu}
In this section we discuss the application of various neural network
based algorithms to the spectral data discussed in Section~\ref{sec:data}.
In order to establish the validity of using the neural network
ensembles to perform this task, we conducted the following sets of 
experiments:
\begin{enumerate}
\item To evaluate the suitability of the neural network methods, we
re-developed constituent algorithms (1) and (2) with the RBF and MLP
ensembles using:
\begin{enumerate}
\item the full-parameter, pre-processed data sets;
\item the diagnostically relevant principal components obtained from the 
full-parameter, pre-processed data sets;
\item the reduced-parameter, pre-processed data sets.
\end{enumerate}
\item Reduce the two-step algorithm developed using MSA into a 
single-step algorithm using RBF and/or MLP ensembles.
\item Produce full comparative results showing the neural
network ensembles' classification accuracy and reliability
relative to the MSA, Pap smear and colposcopy.
\end{enumerate}

\subsection{Neural Network Ensembles on Full-Parameter Set}
\label{sec:respca}
The first step in applying the neural network ensembles to this problem 
consisted of determining whether the algorithms were suited for this
task. To that end, both the MLP and RBF ensembles were used to separate
the normal squamous tissues from the SILs (constituent algorithm (1)).
The task proved to be impossible using the full 160 parameters mainly
because of the small number of training samples. Constraining the
number of weights required to handle 160 parameters with the 
available data resulted in a highly ill-posed problem~\cite{wahb82}.
As a result the network tried to memorize the training data and
performed poorly on the test set. In order to avoid this pitfall,
we used the three  diagnostically relevant principal components from the 
full-parameter data set containing normalized fluorescence spectra.
(Note, the MSA was developed using these same three PCs, because it also
was unable to solve this problem using the full-parameter data.)
Both networks had two outputs, each representing the posterior 
probabilities of the corresponding class.
The MLP networks had a single hidden layer with three hidden units, and
the RBF networks had three kernels which were initialized by a $k$-means
algorithm on the training set\footnote{The appropriate sizes of both the MLP
and the RBF network were determined experimentally. Because we found that the 
performance was comparable over fairly large range of network sizes, it was 
not necessary to use sophisticated methods.}.

The ensemble results reported in Table~\ref{tab:1pca} correspond
to the pooling of $20$ different classifiers (MLPs or RBFs), 
each of which started
from a different random weight initializations, using the
average and median combiners, respectively\footnote{The gains
due to combining were minimal if more classifiers were used.}.
All the results reported in this article are based on test set
performance.
We report the sensitivity and specificity values separately, rather than 
a single classification rate, to emphasize the difference between 
a false-positive and a false-negative.
For an application such as pre-cancer detection, the cost
of a misclassification varies greatly from one class to another.
Erroneously labeling a healthy tissue as pre-cancerous can be
corrected when further tests are performed. Labeling a pre-cancerous
tissue as healthy however, can lead to disastrous consequences.
Therefore, for algorithm (1), we increased the cost of a
misclassified SIL until the sensitivity\footnote{In the results reported
in this section, the cost of misclassifying a SIL was two times the cost of
misclassifying a normal tissue sample.} reached a satisfactory level
(comparable to the sensitivity of current clinical detection).

For this experiment, the RBF based combiners provide higher
specificity than either the MLP ensembles or the MSA, for a
similar sensitivity.
This experiment was conducted to ensure that neural network ensembles
could duplicate the MSA results ~\cite{rami96}, using the principal components
extracted from the full-parameter set.

\begin{table}[ht]
\caption{Accuracy   of {\em constituent} algorithm (1) for
differentiating between SILs and normal squamous  tissues, based
on the diagnostically relevant principal components of the
full-parameter set containing normalized spectra.
All results are based on test test performance.}
\label{tab:1pca}
\begin{center}
\begin{tabular}{|c||cc|} \hline \hline
Algorithm         & Specificity         & Sensitivity \\ \hline
MSA       & 68\%  & 88\%\\
RBF-single& 71\% $\pm 3.8\%$ & 86\% $\pm 2.1\%$ \\
MLP-single& 65\% $\pm 0.0\%$ & 83\% $\pm 1.1\%$ \\ \hline
RBF-ave   & 74\% $\pm 0.7\%$ & 86\% $\pm 0.5\%$ \\
RBF-med   & 72\% $\pm 1.7\%$ & 86\% $\pm 1.5\%$ \\ \hline
MLP-ave   & 65\% $\pm 0.0\%$ & 83\% $\pm 0.0\%$ \\
MLP-med   & 65\% $\pm 0.0\%$ & 84\% $\pm 0.7\%$ \\ \hline
\end{tabular}
\end{center}
\end{table}

\subsection{Neural Networks on Reduced-Parameter Set (2-step Algorithm)}
\label{sec:resred}
The second step in applying neural networks to this problem consisted of
determining whether the neural network ensemble performance
on the reduced-parameter data set was acceptable, by
retracing the development of the two-step process outlined for 
the multivariate statistical algorithm. 
Specifically, the neural network ensemble was applied to the pre-processed
reduced-parameter data sets, used to develop constituent algorithms
(1) and (2).

For {\em constituent} 
algorithm (1) the  RBF kernels were initialized using a $k$-means
clustering algorithm on the training set containing normal squamous tissue
samples and SILs. 
The RBF networks had 10 kernels, whose locations and spreads were adjusted
during training. The MLP had one hidden layer with 10 hidden units.
For {\em constituent} algorithm (2), we selected 10 kernels, half of which
were fixed to patterns from the columnar normal class, while the other half
were initialized using a $k$-means algorithm. Neither the kernel
locations nor their spreads were adjusted during training.
This process was adopted to rectify the large discrepancy between the 
number of 
samples from each category (13 for columnar normal vs. 58 for SILs).
In this case, the MLP algorithm had 10 hidden units.
Training was shypothesizetopped when the performance on the training set
slowed down sufficiently to suggest further training would cause
overtraining\footnote{In general, Leave-One-Out or K-fold cross
validation woudl be more desirable. However, because the results 
were similar over a large window of stopping times, this more naive
method seemed adequate.}

Once a stopping time was selected, 20 cases
were run for each algorithm\footnote{Each run
has a different initialial set of kernels/spreads/weights.}.

\begin{table}[ht]
\caption{Accuracy   of {\em constituent} algorithm (1) for
differentiating  SILs and normal squamous  tissues, based
on the reduced-parameter set containing normalized spectra.}
\label{tab:res_1red}
\begin{center}
\begin{tabular}{|c||cc|} \hline \hline
Algorithm        & Specificity         & Sensitivity \\ \hline
MSA       & 63\% & 90\%   \\ 
RBF-single& 66\% $\pm 2.7\%$ & 84\% $\pm 2.0\%$ \\
MLP-single& 61\% $\pm 0.0\%$ & 91\% $\pm 0.2\%$ \\ \hline
RBF-ave   & 66\% $\pm 0.6\%$ & 90\% $\pm 0.0\%$ \\ 
RBF-med   & 66\% $\pm 1.1\%$ & 90\% $\pm 0.7\%$ \\ \hline
MLP-ave   & 61\% $\pm 0.0\%$ & 91\% $\pm 0.0\%$ \\ 
MLP-med   & 61\% $\pm 0.0\%$ & 91\% $\pm 0.0\%$ \\ \hline
\end{tabular}
\end{center}
\end{table}

\begin{table}[ht]
\caption{Accuracy   of {\em constituent} algorithm (2) for
differentiating  SILs and normal columnar  tissues, based
on the reduced-parameter set containing normalized and
mean-scaled spectra.}
\label{tab:res_2red}
\begin{center}
\begin{tabular}{|c||cc|} \hline \hline
Algorithm &  Specificity &   Sensitivity  \\ \hline
MSA       & 36\%   &   97\% \\
RBF-single& 35\% $\pm 15\%$  & 97\% $\pm 1.6\%$ \\
MLP-single& 47\% $\pm 6.9\%$ & 89\% $\pm 3.5\%$ \\ \hline
RBF-ave   & 37\% $\pm 5.0\%$ & 97\% $\pm 0.0\%$ \\
RBF-med   & 44\% $\pm 7.5\%$ & 97\% $\pm 0.0\%$ \\ \hline 
MLP-ave   & 50\% $\pm 0.0\%$ & 88\% $\pm 0.7\%$ \\
MLP-med   & 50\% $\pm 0.0\%$ & 89\% $\pm 2.5\%$ \\ \hline 
\end{tabular}
\end{center}
\end{table}

The ensemble results reported are based on the pooling of $20$ different
runs of RBF networks, initialized and trained as described in the
previous section. 
Once again we increased the cost of misclassifying a SIL in order
to increase the sensitivity at the expense of reducing the specificity.
For algorithm (1), the cost of a 
misclassified SIL was $2.5$ times the cost of a misclassified normal
tissue sample.
The sensitivity and specificity values for {\em constituent} algorithm (1) 
based on MSA, MLP and RBF ensembles are provided in 
Table~\ref{tab:res_1red}. 
Table~\ref{tab:res_2red} presents sensitivity and specificity values
for {\em constituent}
algorithm (2) obtained from MSA, and MLP  and RBF 
ensembles. In this case,
there was no need to increase the cost of a misclassified SIL for
the RBF network, because
of the high prominence of SILs in the training set.
For the MLPs however, the cost of normal columnars had to be increased
in order to obtain classification decisions\footnote{When the cost
of a misclassified SIL was the same as the cost of a misclassified
normal columnar, all patterns were classified as SILs by the MLP. Only
when the cost of normal columnars was increased did the MLP 
start to make non-trivial classification decisions.}.

For both algorithms (1) and (2), the RBF based combiners 
provide higher specificity than the MSA, while the MLP ensemble
does so only for algorithm (1). Furthermore in the case of the RBF
algorithm, this increase is achieved without a decrease in the
sensitivity.
The median combiner provides results similar to those of the 
average combiner, except for algorithm (2) where it provides 
better specificity. From these experiments we can conclude that
the RBF network is better suited for this task than the MLP.
While in some other work such as classification of sonar signals
we have found combining MLPs with RBFs to be fruitful~\cite{ghtu96},
in this problem we were surprised
to find that combining MLPs and RBFs always gave worse results 
than combining RBFs alone.
We hypothesize that the fine tuning required to accommodate the varying 
number of class samples poses severe problems for MLPs global computations.
The RBF networks significantly alleviate this problem 
by placing the kernels in the appropriate places. 

We conducted this experiment to not only 
to demonstrate that the reduced-parameter data set does not compromise
the performance of the network ensembles,
but also to compare the RBF and MLP ensembles. 
having established the validity of the reduced-parameter set,
we performed the remaining experiments on the reduced-parameter data
set only. This is a significant step, since it allows us to solely 
depend on the $13$ parameters obtained from the component loadings, 
rather than on the principal components from the original $160$ values.
Furthermore, the results of the first two steps indicate that
the RBF network ensembles not only duplicate and 
surpass the MSA results, but also outperform the MLP algorithms at
every step.
Since the MLP ensembles fail to show any areas where they may be
more desirable than the RBF network ensembles, 
we will restrict the remainder of the experiments to RBF network
ensembles only.

In Sections~ \ref{sec:respca} and \ref{sec:resred},
we used the two constituent algorithms separately,
to discriminate different types of normal tissue from SILs.
In order to obtain the final discrimination between normal tissue and
SILs, {\em constituent} algorithms (1) and (2) need to be
used sequentially. This two step approach, highlighted in 
Figure~\ref{fig:schema}, was specifically developed
for the multivariate statistical analysis, which performed best when
the decision tasks were simplified.
In the next section we present a more direct approach that uses the
strengths of the neural network ensembles to reduce the multi-step
classification scheme to a direct, one-step, process.

\subsection{Single Step Classification using Reduced-Parameter Set}
\label{sec:res1st}
In this section, we examine the potential of RBF ensembles
for separating SILs
from normal tissue samples in a single classification step.
Because the pre-processing
for algorithms (1) and (2) is different\footnote{Normalization vs. 
normalization followed by mean scaling.}, we formed 
26-dimensional inputs by concatenating the two reduced
features sets describes in Section~\ref{sec:redpar}. 
We initialized 10 kernels using a $k$-means
algorithm on a trimmed\footnote{The
trimmed set has the same number of patterns from each class. Thus, it
forces each class to have a similar number of kernels. This set is
used {\em only} for initializing the kernels, {\em not} while 
training.} version of the training set. Without this trimming 
process, few patterns belonging to the columnar normal class
are selected as kernels due to their low prominence in the training
set, resulting in poor performance when such a sample is seen.
During training, the kernel locations and spreads were not adjusted, 
to allow kernels to remain in sections of the input space where 
few patterns are observed. 
The cost of a misclassified SIL was set at 2.5 times the 
cost of a misclassified normal tissue sample, in order to provide the
best sensitivity/specificity pair. The average and median
combiner results are obtained by pooling 20 RBF networks\footnote{This
procedure is repeated 10 times, in order to determine the variability
in the ensembles.}.

\begin{table}[ht]
\caption{One step RBF algorithm compared to multi-step MSA. (Based on
reduced-parameter set.)}
\label{tab:1st}
\begin{center}
\begin{tabular}{|c||cc|} \hline \hline
Algorithm          &  Specificity      &  Sensitivity  \\ \hline
2-step MSA         & 65\%             & 84\%  \\
2-step RBF-ave     & 65\% $\pm$ 2\%    & 87\% $\pm$1\%   \\
2-step RBF-med     & 67\% $\pm$ 2\%    & 87\% $\pm$1\%   \\ \hline
1-step RBF-single  & 65\% $\pm$ 9\%   & 89\% $\pm$2.8\% \\ 
1-step RBF-ave     & 67\% $\pm$ .7\%   & 91\% $\pm$1.5\% \\
1-step RBF-med     & 65.5\% $\pm$ .5\% & 91\% $\pm$1\% \\ \hline 
\end{tabular}
\end{center}
\end{table}

Table~\ref{tab:1st} shows the performance of the
algorithms for a given SIL misclassification cost.
For comparison purposes, the results of the 2-step RBF ensemble
algorithms are also provided. These algorithms perform the
same steps as the MSA, using the results presented in 
Section~\ref{sec:redpar} for each constituent algorithm).

As we discussed above, for an application such as pre-cancer 
detection, it is far more critical to increase the classification 
accuracy of some classes than others, to eliminate certain types
of errors. 
By making the algorithm more compact, the one step algorithm also
makes this trade-off more visible.

\subsubsection{Sensitivity-Specificity Tradeoff}
In this subsection
we detail the interaction between the cost of misclassification and
the variation in the sensitivity/specificity of the RBF ensembles.
Since it may be required to reach a predetermined sensitivity, 
we have varied the cost of misclassifying a SIL to obtain a wide
range of sensitivity/specificity pairs.
Table~\ref{tab:cost} shows the specificity and sensitivity for various
costs of a misclassification.

\begin{table}[ht]
\caption{Effect of misclassification cost on 1-step RBF algorithm.} 
\label{tab:cost}
\begin{center}
\begin{tabular}{|l|c|cc|} \hline \hline
Algorithm          &  Cost of SIL &  Specificity     &  Sensitivity  \\ 
                   & misclassification &             & \\ \hline 
2-step MSA         &   & 65\% & 84\% \\ \hline
RBF-ave            & 1 & 85\% & 61\% \\
RBF-med            &   & 84\% & 61\% \\  \hline
RBF-ave            & 2 & 75\% & 88\% \\
RBF-med            &   & 74\% & 86\% \\ \hline
RBF-ave            &2.5& 67\% & 91\% \\
RBF-med            &   & 66\% & 91\% \\ \hline
RBF-ave            & 3 & 63\% & 93\%  \\
RBF-med            &   & 59\% & 93\%  \\ \hline
RBF-ave            & 4 & 55\% & 95\%  \\
RBF-med            &   & 52\% & 97\%  \\ \hline
RBF-ave            & 5 & 39\% & 97\%  \\
RBF-med            &   & 37\% & 97\%  \\ \hline
\end{tabular}
\end{center}
\end{table}

On observing the performance of the RBF ensembles at various costs
of misclassifying SILs, the improvements over the two-step
MSA algorithm are apparent. If the specificity is required to be 
above $60\%$, for example, using a SIL misclassification cost of three 
improves the sensitivity of the MSA by a significant $10\%$, using
the average RBF ensembles. 
If, on the other hand, a sensitivity 
above $83\%$ is required, using a SIL misclassification cost of
two, provides an improvement of $12\%$ over the specificity of
the MSA.
 
\subsection{Comparative Results}
\label{sec:rescomp}
In the previous sections we discussed how the neural network
ensembles were applied to various stages of this classification
tasks. In this section we compare the final results of the 
RBF network ensembles to the best MSA result and to clinical methods.
The SIL misclassification cost of $2.5$ provided the best
compromise between sensitivity and specificity.
To find the variability of the methods, we
performed the ensemble averaging $10$ times on $20$ different
individual classifiers.

The results of both the two-step and one-step RBF algorithms
and  the  results  of the two-step MSA are compared  to  the
accuracy  of  Pap smear screening and colposcopy  in  expert
hands in Table~\ref{tab:comp}. A comparison of one-step RBF algorithms to
the two-step RBF algorithms indicates that the one-step 
algorithms have similar specificities, but a moderate 
improvement in sensitivity relative to the two-step  algorithms.
Compared to the MSA,  the
one-step  RBF algorithms  have  a  slightly  decreased
specificity,  but a substantially improved  sensitivity.  In
addition  to  the improved sensitivity, the  one  step  RBF
algorithms simplify the decision making process.  A
comparison between the one step RBF algorithms and Pap
smear screening indicates that the RBF algorithms  have
a  nearly 30\%  improvement  in sensitivity with  no  compromise  in
specificity; when compared to colposcopy in expert hands~\cite{fair95}, 
the RBF ensemble  algorithms  maintain the sensitivity  of  expert
colposcopists,  while  improving the specificity  by  almost
20\%. 

\begin{table}[ht]
\caption{One step RBF algorithm compared to multi-step 
MSA \protect\cite{rami96} and clinical methods \protect\cite{fair95} 
for differentiating  SILs and normal tissue samples.}
\label{tab:comp}
\begin{center}
\begin{tabular}{|c||cc|} \hline \hline
Algorithm          &  Specificity      &  Sensitivity  \\ \hline
2-step MSA         & 65\%             & 84\%  \\
RBF-single  & 65\% $\pm$ 9\%   & 89\% $\pm$2.8\% \\ 
RBF-ave            & 67\% $\pm$.75\%  & 91\% $\pm$1.5\% \\ 
RBF-med            & 65.5\% $\pm$.5\% & 91\% $\pm$1\% \\ \hline \hline 
Pap smear (human expert)          & 68\% $\pm$21\%   & 62\% $\pm$23\% \\
Colposcopy   (human expert)      & 48\%$\pm$23 \%   & 94\% $\pm$6\% \\ \hline
\end{tabular}
\end{center}
\end{table}

\begin{figure}[htb]
\centerline{\epsfbox{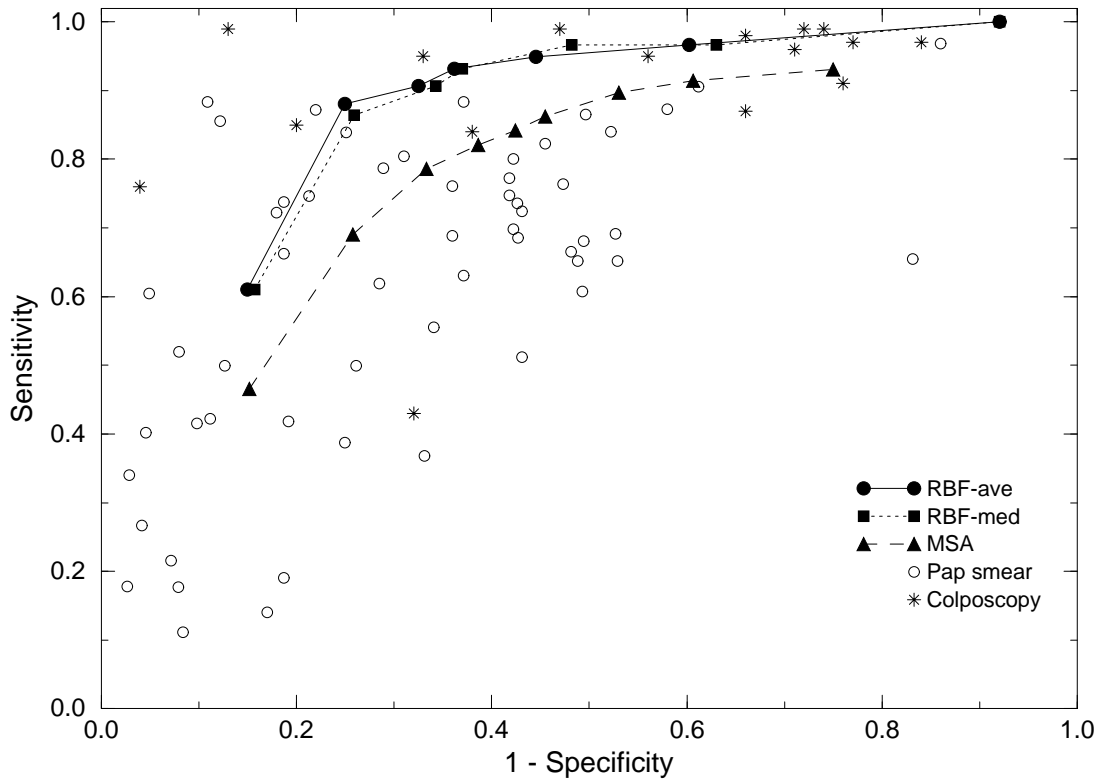}}
  \caption{Trade-off between sensitivity and specifity for MSA 
\protect\cite{rami96} and RBF
ensembles. For reference, Pap smear and colposcopy results from the
literature on various data sets are included \protect\cite{fair95}.}
  \label{fig:sen}
\end{figure}

Figure~\ref{fig:sen} further illustrates the trade-off between specificity
and sensitivity for clinical methods, MSA and RBF ensembles, obtained
by changing the misclassification cost.
The one-step RBF ensembles provide better sensitivity and higher
reliability than any other method for a given specificity value.

\section{Discussion}
\label{sec:disc}
In this article we demonstrate that cervical tissue fluorescence
spectra can be used to develop detection algorithms that differentiate
SILs from normal tissue samples. Of the various algorithms
explored, the RBF network ensemble proved to be the best
alternative, surpassing single networks, MLP ensembles, and the 
multivariate statistical algorithm. 

The  classification results of both  the  multivariate
statistical algorithms and the radial basis function network
ensembles   demonstrate  that  significant  improvement   in
classification  accuracy  can  be  achieved   over   current
clinical detection modalities using cervical tissue spectral
data obtained from {\em in vivo} fluorescence spectroscopy.
The one-step RBF  algorithm  has  the
potential  to  significantly  reduce  the  number  of   
pre-cancerous cases missed by Pap smear screening and the number
of normal tissues misdiagnosed by expert colposcopists. 

The  qualitative nature of current clinical  detection
modalities leads to a significant variability in
classification  accuracy.  For  example,  estimates  of  the
sensitivity and specificity of Pap smear screening have been
shown  to  range from 11-99\% and 14-97\%, respectively~\cite{fair95}.  
This limitation  can  be  addressed by exploiting the variance
reducing properties of an ensemble approach. In particular, RBF 
ensembles demonstrate  a significantly smaller  variability  in
classification accuracy, therefore enabling more reliable
classification.
In  addition  to demonstrating a superior sensitivity,
the  RBF ensembles simplify  the  decision  making
process of the two-step algorithms based on RBF and MSA into
a  single  step that discriminates between SILs  and  normal
tissues. We note that for the given data set, both MSA
and MLP were unable to provide satisfactory solutions in one step.

The one-step algorithm development process can  be
readily   implemented   in  software,   enabling   automated
detection of cervical pre-cancer. It can potentially provide
near real time implementation of pre-cancer detection in
the hands of non-experts, and could lead to wide-scale
implementation of screening and diagnosis, and more effective
patient management in the prevention of cervical cancer. The
success of this application will represent an important step
forward  in  both medical laser spectroscopy and gynecologic
oncology.

\nocite{tugh96b}

{\noindent \bf  Acknoledgements}:
This research was supported in part by NSF
grant ECS 9307632, AFOSR contract F49620-93-1-0307, and Lifespex, Inc.   


\end{document}